\documentclass[10pt,twocolumn,letterpaper]{article} 
\usepackage[pagenumbers]{wacv}

\usepackage{etex}
\usepackage{morefloats}
\usepackage{graphicx} 
\usepackage{morefloats}
\usepackage{amsmath}
\usepackage{amssymb}
\usepackage{booktabs}
\usepackage{multicol}
\usepackage{multirow}
\usepackage{float}
\usepackage[pagebackref,breaklinks,colorlinks]{hyperref}

\usepackage[capitalize]{cleveref}
\crefname{section}{Sec.}{Secs.}
\Crefname{section}{Section}{Sections}
\Crefname{table}{Table}{Tables}
\crefname{table}{Tab.}{Tabs.}

\def\wacvPaperID{19}
\def\confName{WACV}
\def\confYear{2025}

\title{FoundPAD: Foundation Models Reloaded for Face Presentation Attack Detection}

\author{Guray Ozgur$^{1}$, Eduarda Caldeira$^{1}$, Tahar Chettaoui$^{1}$, Fadi Boutros$^{1}$, \\Raghavendra Ramachandra$^{2}$, Naser Damer$^{1,3}$ \vspace{2mm} \\ 
$^{1}$ Fraunhofer IGD, Darmstadt, Germany\\
$^{2}$ Norwegian University of Science and Technology (NTNU), Gjøvik, Norway\\
$^{3}$ TU Darmstadt, Darmstadt, Germany\\
{\small \textcolor{black}{Corresponding author: \hypersetup{urlcolor=black}\href{mailto:guray.ozgur@igd.fraunhofer.de}{guray.ozgur@igd.fraunhofer.de}}}
}
\begin{document}

\maketitle

\begin{abstract}
\vspace{-4mm}
   Although face recognition systems have seen a massive performance enhancement in recent years, they are still targeted by threats such as presentation attacks, leading to the need for generalizable presentation attack detection (PAD) algorithms. Current PAD solutions suffer from two main problems: low generalization to unknown scenarios and large training data requirements. Foundation models (FM) are pre-trained on extensive datasets, achieving remarkable results when generalizing to unseen domains and allowing for efficient task-specific adaption even when little training data are available. In this work, we recognize the potential of FMs to address common PAD problems and tackle the PAD task with an adapted FM for the first time. The FM under consideration is adapted with LoRA weights while simultaneously training a classification header. The resultant architecture, FoundPAD, is highly generalizable to unseen domains, achieving competitive results in several settings under different data availability scenarios and even when using synthetic training data. To encourage reproducibility and facilitate further research in PAD, we publicly release the implementation of FoundPAD at \begin{small}
       \url{https://github.com/gurayozgur/FoundPAD}
   \end{small}.
\end{abstract}

\vspace{-7.5mm}
\section{Introduction}
\vspace{-2mm}
Face recognition (FR) systems have seen a massive performance improvement in recent years, mainly due to the development of deep learning \cite{boutros2022elasticface, deng2019arcface}.
However, they still suffer from different malicious attacks \cite{DBLP:journals/access/MakrushinUD23}.
Presentation attacks (PA) constitute an example of such threats, as they allow the attacker to claim an identity different from their own through techniques such as 3D masks, printed images and replayed videos \cite{casia_fas, oulu_npu, DBLP:conf/eccv/ZhangYLYYSL20, DBLP:journals/pr/FangDKK22}, among others. When left undetected, these attacks can lead to several dangerous situations, such as identity theft \cite{DBLP:journals/access/MakrushinUD23} or unauthorized access to confidential information, as they make the attacker able to impersonate another identity. To tackle this issue, several presentation attack detection (PAD) systems have been proposed \cite{Fang_2022_WACV,DBLP:journals/pami/YuWQLLZ21,DBLP:journals/tbbis/YuLSXZ21,DBLP:conf/cvpr/LiuJ018, DBLP:journals/tcsv/YanZH22,DBLP:journals/tbbis/WangWDG22,DBLP:conf/cvpr/ShaoLLY19,DBLP:conf/cvpr/JiaZSC20,DBLP:conf/aaai/ShaoLY20, fang2024face}. These systems %are designed to 
classify face images as unaltered samples (also known as bona-fide samples) or PAs, allowing the detection of attacks before they are used to verify identities in critical processes. Although these systems can achieve high performance in the intra-dataset scenario \cite{Fang_2022_WACV,DBLP:journals/pami/YuWQLLZ21,DBLP:journals/tbbis/YuLSXZ21,DBLP:conf/cvpr/LiuJ018}, the high sample variability in the cross-domain scenario results in a significant domain change, leading to the need for developing techniques that specifically address this problem \cite{DBLP:journals/tcsv/YanZH22,DBLP:journals/tbbis/WangWDG22,DBLP:conf/cvpr/ShaoLLY19,DBLP:conf/cvpr/JiaZSC20,DBLP:conf/aaai/ShaoLY20, fang2024face}. Another challenge faced by PAD systems is the need for large training datasets to achieve good performance levels \cite{DBLP:conf/eccv/ZhangYLYYSL20, DBLP:conf/cvpr/FangHD23}. 
The performance of PAD networks in low data availability scenarios where smaller training datasets are usually limited to a reduced number of data distributions generally results in less generalizable models \cite{fang2024face}.

Foundation models (FM) contain a large number of trainable parameters that have been pre-trained with self-supervised learning. This learning paradigm allows FMs to learn from unlabeled data, which lifts some common constraints usually associated with model training, namely the difficulty in creating large-scale training datasets. Hence, FMs are trained on extensive and diverse datasets, which results in a generalizable model for a wide range of tasks \cite{DBLP:journals/corr/abs-2108-07258}. Although initially FMs were mainly deployed in natural language processing (NLP) tasks, they have also been effectively used to address computer vision tasks \cite{kirillov2023segment, ravi2024sam, oquab2023dinov2, radford2021learning}. In biometrics, FM utilization is still in an early exploratory stage, with very recent works addressing tasks such as synthetic face image generation \cite{papantoniou2024arc2face}, iris segmentation \cite{farmanifard2024iris} and FR \cite{chettaoui2024froundation}. Biometrics tasks such as cross-domain PAD are expected to highly benefit from the generalization power of FMs, namely in challenging scenarios with low data availability, where they have shown to outperform models trained from scratch \cite{chettaoui2024froundation}. However, this is still an unexplored path for addressing the PAD task. 

\newpage
This work takes advantage of FMs potential to perform generalizable PAD with low data requirements.  In particular, we adapt the pre-trained FM CLIP \cite{radford2021learning} to the PAD task using a low-rank adaption (LoRA), which allows the network to adapt its feature space to the downstream PAD task without losing the knowledge acquired during its self-supervised pre-training, while simultaneously training a header to perform classification. This adapted FM corresponds to our proposed framework, FoundPAD. To show that FoundPAD is effectively taking advantage of FMs' properties, we further propose and evaluate three alternative FM and transformer-based methods. First, we assess the importance of the FM's pre-trained weights by comparing FoundPAD with models following the same architectures trained from scratch (ViT-FS). Then, we prove the importance of properly adapting the FM to the considered downstream task by assessing CLIP's zero-shot performance on PAD (TI) and evaluating a training scenario where no LoRA adaptation is performed and only the classification layer is trained (FE). With these experiments, we were able to prove FoundPAD's superiority in taking advantage of FMs' potential to perform PAD, with FoundPAD achieving an average single-source HTER lower than the second best-performing method in the literature by 6.54 percentage points (pp.) while surpassing ViT-FS and FE by 4.35 pp. and 8.94 pp., respectively, for CLIP ViT-L. Hence, this work contributes to a better understanding of the generalizability capacities of FMs, through our FoundPAD, proving their capacity to surpass previously existent SOTA PAD solutions in many of the considered benchmarking protocols.

\vspace{-2mm}
\section{Related Work} \label{sec:sota}
\vspace{-2mm}
\paragraph{Cross-Domain PAD:} The recent deep learning advances have boosted the development of high-performing face PAD solutions \cite{Fang_2022_WACV,DBLP:journals/pami/YuWQLLZ21,DBLP:journals/tbbis/YuLSXZ21,DBLP:conf/cvpr/LiuJ018}, especially in intra-dataset evaluation scenarios. However, PAs can take several forms e.g., printed images, replayed videos, and 3D masks, and the samples used for model training can be acquired under different conditions for distinct datasets, resulting in a significant domain shift when assessing the performance of PAD methods under the cross-dataset scenario. This results in degraded performance of the SOTA intra-dataset PAD models, leading to the need for developing techniques that perform well across different domains \cite{DBLP:journals/tcsv/YanZH22,DBLP:journals/tbbis/WangWDG22,DBLP:conf/cvpr/ShaoLLY19,DBLP:conf/cvpr/JiaZSC20,DBLP:conf/aaai/ShaoLY20, fang2024face}. These methods can be grouped into domain adaption (DA) and domain generalization (DG) strategies. DA methods \cite{DBLP:journals/tifs/LiLCWHK18,DBLP:conf/icb/WangHSC19} train PAD networks with both labeled source domain and unlabeled target domain data to learn a discriminative feature space that can be used to perform PAD across different domains efficiently. Since only the labels of the source domain are considered, knowledge is transferred to the target domain by aligning its feature space with the one produced by the source features, using techniques like maximum mean discrepancy minimization \cite{DBLP:journals/tifs/LiLCWHK18} and adversarial training \cite{DBLP:conf/icb/WangHSC19}. Nonetheless, two main problems arise when considering DA-based PAD strategies. First, DA-based PAD requires collecting target domain data, which is often difficult and time-consuming. %even if no label annotations are required. 
Second, even if this data is available, using it during the training process is not representative of real-world scenarios, where the target domains may not be completely known, making access to their data usually inexistent during training \cite{fang2024face}. To avoid the need to rely on target testing data, DG strategies \cite{DBLP:conf/cvpr/ShaoLLY19,DBLP:conf/cvpr/JiaZSC20,DBLP:conf/aaai/ShaoLY20,DBLP:conf/aaai/ChenYSDTLHJ21, fang2024face} use training data from several sources simultaneously to enable a broader understanding of distinct domains by jointly analyzing their data distributions. DG-based PAD methods follow different approaches, namely adversarial training \cite{DBLP:conf/cvpr/ShaoLLY19,DBLP:conf/cvpr/JiaZSC20} and meta-learning \cite{DBLP:conf/aaai/ShaoLY20,DBLP:conf/aaai/ChenYSDTLHJ21}. Although these approaches have achieved good results in unseen target domains, they rely on the availability of labeled data from several sources, which is challenging to satisfy in practice, and sometimes rely upon multi-stage or multi-network training strategies \cite{DBLP:conf/icmcs/LiuCDLZX22,DBLP:conf/aaai/ChenYSDTLHJ21}, which induce high computational costs. 

\vspace{-5mm}
\paragraph{Low Data Availability:} In addition to the cross-domain issues, the necessity of large PAD training datasets has been brought up in several works \cite{DBLP:conf/eccv/ZhangYLYYSL20, DBLP:conf/cvpr/FangHD23}. In particular, the main two available large-scale datasets, CelebA-Spoof \cite{DBLP:conf/eccv/ZhangYLYYSL20} and SynthASpoof \cite{DBLP:conf/cvpr/FangHD23}, crossed the threshold of 2k bonafide and 4k attack samples, resulting in a lack of large-scale datasets that is particularly problematic in cross-domain PAD, where smaller training datasets are usually limited to a small number of domains, resulting in less generalizable models \cite{fang2024face}.

\vspace{-5mm}
\paragraph{Foundation Models:} FMs are built with a vast number of trainable parameters, enabling their training on extensive and diverse datasets, which is particularly beneficial for fields that tackle a wide variety of tasks, such as computer vision \cite{kirillov2023segment, radford2021learning, farmanifard2024iris, oquab2023dinov2, papantoniou2024arc2face, chettaoui2024froundation}. Kirillov~\textit{et al.}\cite{kirillov2023segment} designed the Segment Anything Model (SAM) for image segmentation across various domains, achieving a remarkable generalization capacity that makes it able to handle novel image distributions. DINOv2 networks \cite{oquab2023dinov2} are self-supervised pretrained visual models capable of generating universal features for image-level and pixel-level tasks. Radford~\textit{et al.}~\cite{radford2021learning} introduced Contrastive Language-Image Pretraining (CLIP), a multimodal FM trained to process visual and textual inputs, thus learning the relationships between them.

Although the extensive training data used by vision FMs grants them a high degree of generalizability, their performance often falls short in specialized settings \cite{DBLP:journals/corr/abs-2308-07156}. As a result, several techniques that adapt Vision Transformer (ViT) networks to specific downstream tasks have been proposed \cite{chen2022vision, chen2022adaptformer, hu2021lora}. ViT-Adapter \cite{chen2022vision} achieved SOTA performance on the COCO dataset by incorporating fine-grained multi-scale feature reconstruction and embedding image-specific inductive biases into the FM. AdaptFormer \cite{chen2022adaptformer} replaced the standard MLP block in the transformer encoder with two parallel MLP branches, one that mirrors the original network, preserving its generalization capabilities, and another for task-specific fine-tuning. Hu~\textit{et al.}~\cite{hu2021lora} defined the concept of Low-Rank Adaptation (LoRA) layers, which consist of trainable rank decomposition matrices that are inserted in the pre-trained FM. When new data is fed to this network, its weights are frozen and only the LoRA weights are fine-tuned, allowing to adapt the FM to a new task without disregarding its previously acquired knowledge. LoRA has proven effective across diverse applications, namely capsule endoscopy diagnosis \cite{zhang2024learning}, plant phenotyping \cite{chen2023adapting}, and FR \cite{chettaoui2024froundation}. In this study, we select LoRA to adapt CLIP for the PAD task, given its promising results in biometrics \cite{chettaoui2024froundation}.

% COMMENT: the sentence above is a bit in contradiction with this...
Despite the increasing attention FMs have received in recent years, their application in the biometrics field remains largely underexplored. Papantoniou~\textit{et al.}~\cite{papantoniou2024arc2face} used FMs to synthesize facial images conditioned on identity-specific information. Farmanifard~\textit{et al.}~\cite{farmanifard2024iris} adapted SAM \cite{kirillov2023segment} to perform iris segmentation. A recent work \cite{chettaoui2024froundation} applied LoRA \cite{hu2021lora} to adapt DINOv2 \cite{oquab2023dinov2} and CLIP \cite{radford2021learning} to the FR task. The obtained results showcase the benefits of adapting FMs with LoRA, as the adapted networks showed competitive performances to models trained from scratch, outperforming them in challenging scenarios with low data availability. As will be proved later in Section \ref{sec:results}, FMs can also boost PAD performances in low data availability settings. 

In this work, we recognize that FMs can be particularly helpful in addressing tasks usually associated with generalization problems, such as cross-domain PAD. Simultaneously, FMs may also result in great benefits when applied to the challenging low data availability scenario that commonly affects PAD \cite{chettaoui2024froundation}.

\vspace{-3mm}
\section{Methodology} \label{sec:methodology}

\vspace{-1mm}
\subsection{CLIP: A preliminary} \label{sec:clip}
\vspace{-2mm}
CLIP \cite{radford2021learning} is a multimodal FM designed to process visual and textual inputs simultaneously. In this work, we select it as the base FM to perform PAD due to its capacity to achieve competitive results in zero-shot learning scenarios \cite{radford2021learning} and  generalize across a wide range of tasks \cite{radford2021learning} and to specific downstream tasks when adapted with LoRA \cite{chettaoui2024froundation}, which is of high importance in domain-specific settings as the ones imposed by PAD. CLIP was trained on a vast dataset of paired image and text samples, enabling it to learn the connections between these two modalities. Its architecture employs two separate encoders: one for images and another for text. These encoders are trained simultaneously using a contrastive learning objective that measures the cosine similarity between their output features. For positive pairs (where the text matches the image), the similarity is maximized, while for negative pairs (where they do not correspond), the similarity is minimized. This training strategy enables CLIP to effectively capture the semantic relationships between images and their descriptions, resulting in a highly versatile model capable of generalizing across diverse tasks \cite{radford2021learning}. Furthermore, CLIP demonstrates strong performance in zero-shot learning scenarios, where no task-specific fine-tuning is required.

In this work, and to rationalise our FoundPAD, we leverage CLIP to process both image-text pairs and single-image inputs in various approaches. Initially, we assess CLIP's potential as a PAD solution by evaluating its ability to distinguish between bona-fide and PA samples without additional training, as described in Section \ref{sec:baseline}. Towards that, CLIP processes image-text pairs where the text describes the possible labels of the image (``biometric presentation attack'' and ``bona-fide presentation'' following ISO/IEC 2382-37 \cite{ISO238237}). In other approaches, we use only the image encoder, treating CLIP as a feature extractor. A binary classification layer is added on top of the extracted features to perform PAD, following recent works that successfully applied FMs to downstream tasks such as image segmentation \cite{kirillov2023segment} and FR \cite{chettaoui2024froundation}. In these approaches, later detailed in Sections \ref{sec:foundPAD} and \ref{sec:baseline}, we initialize CLIP with the pre-trained weights made publicly available in \cite{radford2021learning}\footnote{\url{https://github.com/OpenAI/CLIP}}.

\vspace{-2mm}
\subsection{FoundPAD} \label{sec:foundPAD}
\vspace{-2mm}
In this work, we propose to adapt FMs to the downstream task of PAD, taking advantage of their high capacity to generalize to novel domains. To this end, we propose a framework that adapts the pre-trained FM with LoRA layers, FoundPAD, shifting the generated feature space in a direction that facilitates samples' classification as bona-fide or PAs. As discussed in Section \ref{sec:clip}, some FMs, such as CLIP, consist of two main components: a text encoder and an image encoder. This dual architecture allows CLIP to classify images into a specific group of categories, by defining a text input that describes each of them. In the PAD scenario, these textual inputs simultaneously describe the two possible labels, ``biometric presentation attack'' and ``bona-fide presentation'' \cite{ISO238237}. However, FMs can also process images without any textual input, functioning as a feature extraction tool that can be combined with a classification layer for task-specific classification. In this study, we investigate the FMs’ effectiveness as a feature extractor for PAD by removing the text encoder and adapting the image encoder to extract relevant features to perform PAD.

\vspace{-5mm}
\paragraph{Fine-Tuninig with LoRA:} The image encoder of FMs such as CLIP, is composed of alternating multi-headed self-attention (MSA) layers and multi-layer perceptron (MLP) blocks, with layer normalization applied before each block and residual connections following each block. While FMs can be used for PAD out-of-the-box without fine-tuning, this may lead to suboptimal performances since the produced embedding space is not necessarily optimal for PAD, as will be shown in our detailed experiments. However, training large FMs from scratch would probably result in losing their FM properties, as we will also show in Section \ref{sec:results}. Hence, we opt to incorporate a ViT adapter \cite{chen2022adaptformer, chen2022vision, hu2021lora} to fine-tune CLIP, as described in Section \ref{sec:sota}. In particular, we use LoRA \cite{hu2021lora} for this purpose, due to its capacity to adapt FMs to highly specific-domain tasks \cite{zhang2024learning, chen2023adapting} and given its strong performance in related fields, such as FR \cite{chettaoui2024froundation}.

LoRA leverages low-dimensional reparameterization, which has been proven to be as effective as training the full parameter space \cite{aghajanyan2020intrinsic} while significantly reducing the number of trainable parameters. When using this technique, CLIP's original weights are frozen and a set of trainable rank-decomposition matrices are introduced into each layer of the transformer architecture, enabling efficient adaption with minimal parameter updates. Given a pretrained weight matrix $W_0 \in \mathbb{R}^{d \times k}$, the low-rank decomposition introduced by LoRA updates it as: 
\vspace{-1mm}
\begin{equation} \footnotesize
    W_0+ \Delta W=W_0+\gamma_r BA
    \vspace{-1mm}
\end{equation} 
where $B \in \mathbb{R}^{d \times r}$ and $A \in \mathbb{R}^{r \times k}$ are the trainable rank-decomposition matrices, with the rank $r<<min(d,k)$, and $\gamma_r$ is a scaling factor. While $\gamma_r$ was originally defined as $\frac{\alpha}{r}$, the constant $\alpha$ tends to cause gradient collapse as $r$ increases, leading to an absence of performance improvements for higher ranks \cite{kalajdzievski2023rank}, although more trainable parameters are used for fine-tuning. This problem can be fixed by using rank-stabilized LoRA (rsLoRA) \cite{kalajdzievski2023rank}, which scales $BA$ with $\frac{\alpha}{\sqrt{r}}$ instead of $\frac{\alpha}{r}$, allowing higher ranks to perform better due to the absence of gradient collapse. Hence, we opt to use rsLoRA to fine-tune CLIP, by setting $\gamma_r=\frac{\alpha}{\sqrt{r}}$. As previously mentioned, $W_0$ is kept frozen, and, since $\gamma_r$ is constant, only $A$ and $B$ are updated during fine-tuning. After the fine-tuning process is complete, the final model weights, $W$, are computed by adding the original weights to the LoRA weights, $W=W_0+\gamma_r BA$, which does not introduce additional parameters during inference, preserving the computational efficiency of the original model while benefiting from the fine-tuning improvements.

To ensure efficiency and reduce the number of parameters in the proposed approach, LoRA is applied only to the MSA weights, leaving the MLP blocks unchanged \cite{hu2021lora}. While LoRA can be applied to the projection matrices of the query, key, value, and output ($q$, $k$, $v$, and $o$, respectively) in the MSA, we limit the adaption to the $q$ and $v$ matrices. This choice follows the recommendations from the original LoRA paper \cite{hu2021lora} and a recent application based on FMs in the biometrics field \cite{chettaoui2024froundation}. The MSA mechanism involves $h$ parallel attention heads, each with its own set of $q$, $k$, and $v$ matrices. Each head is adapted independently with LoRA, resulting in distinct LoRA weights for each of them. When an embedding $x$ is processed through the MSA, the projections $Q_i$, $K_i$, and $V_i$ (for the query, key, and vector, respectively) in head $i$ are determined as follows:
\vspace{-1mm}
\begin{equation} \footnotesize
    \label{eq:qkv_projection}
    \begin{gathered}
        Q_i=W_i^qx+\gamma_r B_i^qA_i^qx\\
        K_i=W_i^kx\\
        V_i=W_i^vx+\gamma_r B_i^vA_i^vx
    \end{gathered}
    \vspace{-1mm}
\end{equation}
where $W_i^q$, $W_i^k$ and $W_i^v$ are the frozen projection layers for $q$, $k$ and $v$, respectively, and $A_i^q$, $B_i^q$, $A_i^v$ and $B_i^v$ correspond to the trainable LoRA layers' parameters. The attention score for head $i$, denoted as $Attention(Q_i, K_i, V_i)$, can then be computed as shown in Equation \ref{eq:attention}:
\vspace{-2mm}
\begin{equation} \footnotesize
    \label{eq:attention}
    Attention(Q_i,K_i,V_i)=Softmax\Big(\frac{Q_iK_i^T}{\sqrt{d_k}}\Big)V_i
    \vspace{-1mm}
\end{equation}
where the scaling factor $d_k$ represents the dimension of the key vectors. The MSA layer's output is generated by the projection layer $O$, which takes the concatenated attention scores from all heads along the feature axis as input:
\vspace{-1mm}
\begin{equation}\footnotesize
    \label{eq:multihead}
    Multihead(Q,V,K)=Concat(head_1, ..., head_k)W^0
    \vspace{-1.5mm}
\end{equation}

The output from the MSA is then passed through the static MLP, completing the processing within a single ViT block. The output from ViT block $l$ is subsequently fed into block $l+1$, which consists of a new MSA fine-tuned with LoRA, followed by another fixed MLP. 

\vspace{-5mm}
\paragraph{Classification:} Using LoRA to fine-tune the FMs results in a feature space adapted to the PAD task and, thus, is expected to yield better results. The classification is performed by an extra fully connected layer responsible for processing the features extracted by the adapted FM and output the final predictions, $\tilde{y}$. Finally, the binary cross-entropy loss is used to compare $\tilde{y}$ with the ground truth labels, $y$, allowing to update the model's trainable parameters:
\vspace{-3mm}
\begin{equation} \footnotesize
    \label{eq:bce}
    L_{BCE}=-(y\, log(\tilde{y})+(1-y)\,log(1-\tilde{y}))
        \vspace{-1mm}
\end{equation}

During testing, FoundPAD's weights are frozen and the highest output score produced by the classification layer's neurons defines the model prediction for each sample.

\vspace{-2mm}
\subsection{Baselines} \label{sec:baseline}
\vspace{-2mm}
To analyze the usage of FMs to perform PAD in detail and prove the effectiveness of FoundPAD when compared with alternative baseline solutions, we present three alternative FM or transformer-based methods for PAD.
\textbf{Text-Image (TI):} FMs, such as CLIP, have demonstrated remarkable performance in zero-shot learning scenarios across various downstream tasks, including food classification, car model classification, and offensive memes identification \cite{radford2021learning}. These tasks involve the simultaneous use of text and image encoders for classification. To explore the zero-shot learning capabilities of the selected FM in the PAD task, we evaluated its performance using image-text pairs where the text describes the possible labels of the image. Specifically, each test sample was paired with textual descriptions of its possible labels: ``biometric presentation attack'' and ``bona-fide presentation''. The similarity score between the image embedding and the corresponding text embeddings determined the predicted label. This approach utilizes the complete FM architecture without requiring further training. However, due to the lack of task-specific adaptation and the domain-specific nature of PAD, the TI approach is anticipated to perform suboptimally compared to alternative scenarios or more general visual tasks (e.g. detecting if an image contains a cow).
\textbf{ViT Trained from Scratch (ViT-FS):} Part of the FMs' abilities is related to the architectures they are based on (commonly ViT architectures \cite{alexey2020image}), which have shown promising results when trained for both PAD \cite{huang2022adaptive} and morphing attack detection \cite{zhang2024generalized}. Hence, we investigate if ViT networks can individually contribute to building a strong PAD, by training from scratch identical architectures to the ones used in the considered FMs (ViT-B and ViT-L), using only the selected PAD training dataset. Since the transformer parameters are randomly initialized, ViT-FS cannot be considered an FM, as it does not benefit from any knowledge acquired during any previous training. This allows for direct comparison between FM-based techniques such as FoundPAD and visual transformers, making it possible to assess how valuable the in-built knowledge of FMs is for downstream tasks such as PAD.
\textbf{Feature Extractor (FE):} To evaluate the impact of the network adaptation enabled by LoRA in FoundPAD, we design an experiment in which the FM is frozen without adaption and used as a feature extractor. PAD is then performed by a binary classification layer trained on top of the FM's feature space following Equation \ref{eq:bce}. This experiment allows us to evaluate if the FM's original feature space captures relevant information to perform PAD while providing a reference for quantitatively measuring the improvements resulting from adapting the model's weights with LoRA.

\vspace{-3mm}
\section{Experimental Setup} \label{sec:exp_setup}
\vspace{-1mm}
\textbf{Datasets:} To allow for a wide range of fair comparisons, experiments were conducted on five publicly available datasets widely used in cross-dataset PAD works to benchmark their performances \cite{DBLP:conf/icmcs/LiuCDLZX22,Fang_2022_WACV,DBLP:conf/cvpr/LiPWK18,DBLP:conf/cvpr/ShaoLLY19,DBLP:conf/aaai/ShaoLY20,DBLP:conf/fgr/PanwarSSPG21, fang2024face}: MSU-MFSD \cite{msu_mfs} (denoted as M), CASIA-FASD \cite{casia_fas} (denoted as C), Idiap Replay-Attack \cite{replay_attack} (denoted as I), OULU-NPU \cite{oulu_npu} (denoted as O), and CelebA-Spoof \cite{DBLP:conf/eccv/ZhangYLYYSL20} (denoted as CA). Given the adaption of synthetic data as a privacy-friendly alternative for authentic data in biometric development \cite{DBLP:conf/cvpr/FangHD23}, we additionally use the synthetic-based face PAD dataset SynthASpoof \cite{DBLP:conf/cvpr/FangHD23} as a training dataset paired with the mentioned four datasets as evaluation benchmarks, following the protocol defined in \cite{DBLP:conf/cvpr/FangHD23}. Apart from bonafide samples, the \textbf{MSU-MFSD} \cite{msu_mfs} dataset includes printed photo and replay attacks, totalling 440 videos from 35 subjects. The \textbf{CASIA-FASD} \cite{casia_fas} includes 600 videos from 50 subjects and contains warped photo, cut photo and video replay attacks. The \textbf{Idiap Replay-Attack} \cite{replay_attack} dataset consists of 300 videos from 50 subjects and includes both print and replay attacks. The \textbf{OULU-NPU} \cite{oulu_npu} is a mobile face PAD dataset and contains 5940 videos from 55 subjects, acquired with six distinct mobile phones. The \textbf{CelebA-Spoof} \cite{DBLP:conf/eccv/ZhangYLYYSL20} is diverse in terms of subjects, illumination, and  sensors and comprises four types of attacks, namely print, replay, 3D mask and paper cut attacks. Its images were collected from the web, resulting in a large-scale dataset with 625,537 images from 10,177 subjects. The \textbf{SynthASpoof} \cite{DBLP:conf/cvpr/FangHD23} dataset is designed to address privacy and scalability challenges in PAD research. It contains 25,000 subjects each having only one sample generated using StyleGAN2-ADA \cite{StyleGAN2-ADA}, filtered through CR-FIQA \cite{boutros2023crfiqafaceimagequality} to ensure high quality and realistic appearance. For print attacks, 3,800 subjects were printed and recaptured. Replay attacks were recorded by displaying synthetic images on two different screens and recapturing using three devices, where each device contributed 25,000 images adding up to a total of 75,000 attack images.

All the performed experiments target the cross-dataset scenario, meaning that samples from different datasets are used during training and testing. The number of datasets used for training should be taken into account, as models trained on data from different datasets can learn from distinct information sources and, thus, are expected to perform better in cross-dataset scenarios. Hence, the developed experiments are divided into three groups based on the scale of the data available for training and following established evaluation protocols: triple-source (3 training datasets), double-source (2 training datasets) and single-source (1 training dataset). We perform five triple-source experiments (training dataset(s) → testing dataset): O\&C\&I → M, O\&M\&I → C, O\&C\&M → I, I\&C\&M → O, O\&C\&M → CA,  following previous works \cite{DBLP:conf/aaai/ChenYSDTLHJ21,DBLP:conf/icmcs/LiuCDLZX22,DBLP:conf/cvpr/JiaZSC20,DBLP:conf/cvpr/Wang0SC20,fang2024face}. For the double-source scneario, two cases are considered: M\&I → C and M\&I → O \cite{DBLP:journals/tbbis/YuLSXZ21,DBLP:journals/pami/YuWQLLZ21,Fang_2022_WACV,DBLP:conf/icip/MingYAVLB22, fang2024face}. The single-source scenario includes a set of twelve experiments where one of the M, C, I, and O datasets is used to train the network and the remaining three are separately used for testing, following previous works on cross-dataset PAD \cite{DBLP:journals/tifs/WangHSC21,DBLP:conf/cvpr/Wang0SC20,DBLP:conf/aaai/ShaoLY20,DBLP:conf/cvpr/JiaZSC20,fang2024face}. The SynthASpoof dataset is used to adapt models from the synthetic domain to the authentic domain. Following Meiling~\textit{et al.}~\cite{DBLP:conf/cvpr/FangHD23}, models are trained on only the SynthASpoof dataset and then evaluated on M, C, I, and O.
\textbf{Image Pre-Processing:} 
Before being fed into the image encoder, the training images undergo a preprocessing procedure.

We detect the face in each sample using MTCNN \cite{MTCNN} and resize it to $256\times256$ pixels, following \cite{fang2024face}. During training, all samples are also subject to the data augmentation process later defined in 
this section. Then, the images are processed in a way that allows them to be tokenized, given the success achieved with tokenization in FMs' NLP applications \cite{alexey2020image}. The tokenization process follows the procedure proposed by Alexey~\textit{et al.}~\cite{alexey2020image}. First, the image is divided into non-overlapping regions, which are then processed by a linear projection layer to create patch embeddings. These patch embeddings are combined with a learnable class (CLS) token \cite{alexey2020image}, forming a unified representation of the image that aids in classification. Position embeddings are also added to preserve the spatial order of the original image patches. The resulting embedding vector, enriched with patch-level information, position data, and the CLS token, is then used as the input for CLIP's image encoder \cite{alexey2020image}.
\textbf{Model Architecture:} The used FM, CLIP \cite{radford2021learning}, introduced four models based on two distinct architectures: base and large. The base architecture, comprising 86M parameters, is available in two versions differing by patch size (16 and 32). In contrast, the large architecture has 0.3 billion parameters and includes a variant fine-tuned at a higher resolution of 336 pixels for one extra epoch to enhance performance \cite{touvron2019fixing}. Inspired by recent studies leveraging CLIP for FR \cite{chettaoui2024froundation}, we select one model from each architecture for evaluation: the base model with a patch size of 16 and the large model trained without higher-resolution inputs, from now referred to as ViT-B and ViT-L, respectively. These architectures were used in all our experiments, whether under the TI, ViT-FS, FE, or FoundPAD settings. 
\textbf{Implementation Details:} All of the models presented in this study were trained for 40 epochs using the AdamW \cite{loshchilov2018decoupled} optimizer with a momentum of 0.9 and a weight decay of 0.05, following FRoundation \cite{chettaoui2024froundation}. The batch size was set to 512 for all experiments \cite{chettaoui2024froundation}, except for ViT-L-FS, where a batch size of 432 was used due to GPU constraints. For FoundPAD, the LoRA $r$, $\alpha$ and dropout were set to 8, 8 and 0.4, respectively. For ViT-FS and FoundPAD, the learning rate of the ViT network was defined as 1e-6. The learning rate of the binary classification layer was set to 1e-3 for all experiments requiring training (ViT-FS, FE, and FoundPAD). The data was augmented using random crop to $224\times224$ pixels, random horizontal flip, random gamma correction with gamma limits 80 and 180, an RGB shift with a limit of 20 for each colour component and colour jitter (with brightness, contrast, saturation and hue set to 0.1), following \cite{fang2024face}.
\textbf{Evaluation Metrics:} Following previous work on cross-domain PAD \cite{DBLP:conf/icmcs/LiuCDLZX22,Fang_2022_WACV,DBLP:conf/cvpr/LiPWK18,DBLP:conf/cvpr/ShaoLLY19,DBLP:conf/aaai/ShaoLY20,fang2024face}, the Half Total Error Rate (HTER) and the Area under the Receiver Operating Characteristic (ROC) Curve (AUC) value are determined in percentage (\%), for all performed experiments. HTER is defined as the mean of the standard PAD evaluation metrics Bona fide Presentation Classification Error Rate (BPCER) \cite{ISO301073} and Attack Presentation Classification Error Rate (APCER) \cite{ISO301073}.

\vspace{-3mm}
\section{Results and Discussion} \label{sec:results}

\begin{table*}[h]
\tiny
\vspace{-3mm}
\begin{center}
\caption{Zero-shot learning (TI) results of the considered FM architectures, CLIP ViT-B and ViT-L, on five evaluation benchmarks. Both architectures present very limited classification capabilities, achieving close to random performance for most of the analyzed scenarios.
}
\vspace{-3mm}
\resizebox{0.9\textwidth}{!}{
\begin{tabular}{l|cc|cc|cc|cc|cc}
\hline
\multirow{2}{*}{Model} & \multicolumn{2}{c|}{M} & \multicolumn{2}{c|}{C} & \multicolumn{2}{c|}{I} & \multicolumn{2}{c|}{O} & \multicolumn{2}{c}{CA}\\
 & HTER(\%) $\downarrow$ & AUC(\%) $\uparrow$ & HTER(\%) $\downarrow$ & AUC(\%) $\uparrow$ & HTER(\%) $\downarrow$ & AUC(\%) $\uparrow$ & HTER(\%) $\downarrow$ & AUC(\%) $\uparrow$ & HTER(\%) $\downarrow$ & AUC(\%)  $\uparrow$   \\
\hline \hline
ViT-B & 55.71 & 41.22 & 50.67 & 49.53 & 50.50 & 50.74 & 52.05 & 47.87 & 56.07 & 42.02 \\ 
ViT-L & 41.19 & 62.96 & 43.44 & 56.56 & 46.50 & 54.49 & 44.76 & 59.44 & 58.07 & 39.39 \\ \hline
\end{tabular}}
\label{tab:no_training}
\end{center}
\vspace{-3mm}
\end{table*}

\begin{table*}[]
\tiny
\vspace{-3mm}
\begin{center}
\caption{Results of triple-source cross-dataset evaluation on four benchmarking datasets. The best and second-best results for each metric are highlighted in bold and underlined, respectively. It can be seen that FoundPAD surpasses both ViT-FS and FE in most of the evaluation scenarios while achieving comparable or even superior performances than SOTA PAD methods.}
\vspace{-3mm}
\resizebox{0.95\textwidth}{!}{
\begin{tabular}{ll|cc|cc|cc|cc||cc}
\hline
\multicolumn{2}{c|}{\multirow{2}{*}{Method}} & \multicolumn{2}{c|}{O\&C\&I → M} & \multicolumn{2}{c|}{O\&M\&I → C} & \multicolumn{2}{c|}{O\&C\&M → I} & \multicolumn{2}{c||}{I\&C\&M → O} & \multicolumn{2}{c}{Average} \\ %\cline{2-9} 
 & & HTER(\%) $\downarrow$ & AUC(\%) $\uparrow$ & HTER(\%) $\downarrow$ & AUC(\%) $\uparrow$  & HTER(\%) $\downarrow$ & AUC(\%) $\uparrow$ & HTER(\%) $\downarrow$ & AUC(\%) $\uparrow$  & HTER(\%) $\downarrow$ & AUC(\%) $\uparrow$  \\ \hline \hline
\multicolumn{2}{c|}{Binary CNN  \cite{DBLP:conf/eccv/XuLNX14}}               & 29.25       & 82.87       & 34.88       & 71.94       & 34.47       & 65.88       & 29.61       & 77.54       & 32.05        & 74.56        \\
\multicolumn{2}{c|}{Auxiliary  \cite{DBLP:conf/cvpr/LiuJ018}} & 22.72       & 85.88       & 33.52       & 73.15       & 29.14       & 71.69       & 30.17       & 77.61       & 28.89        & 77.08        \\
\multicolumn{2}{c|}{ResNet50-PS \cite{DBLP:journals/tbbis/YuLSXZ21}} & 14.32       & 94.51       & 18.23       & 89.75       & 18.86       & 89.63       & 21.44       & 87.56       & 18.21        & 90.36        \\
\multicolumn{2}{c|}{NAS-FAS \cite{DBLP:journals/pami/YuWQLLZ21}}   & 19.53       & 88.63       & 16.54       & 90.18       & 14.51       & 93.84       & 13.8        & 93.43       & 16.10        & 91.52        \\
\multicolumn{2}{c|}{LMFD \cite{Fang_2022_WACV}}  & 10.48       & 94.55       & 12.50       & 94.17       & 18.49       & 84.72       & 12.41       & 94.95       & 13.47        & 92.10        \\
\multicolumn{2}{c|}{ViTransPAD \cite{DBLP:conf/icip/MingYAVLB22}} & 8.39        & -         & 21.27       & -         & 16.83       & -         & 15.63       & -          & 15.53        & -            \\
\multicolumn{2}{c|}{PatchNet \cite{DBLP:conf/cvpr/WangLYL22}}  & 7.10        & \textbf{98.64}       & 11.33       & 94.58       & 14.60       & 92.51       & 11.82       & 95.07       & 11.21        & 95.20        \\
\multicolumn{2}{c|}{MADDG \cite{DBLP:conf/cvpr/ShaoLLY19}}        & 17.69       & 88.06       & 24.50       & 84.51       & 22.19       & 84.99       & 27.89       & 80.02       & 23.07        & 84.40        \\
\multicolumn{2}{c|}{RFM \cite{DBLP:conf/aaai/ShaoLY20}}        & 17.30       & 90.48       & 13.89       & 93.98       & 20.27       & 88.16       & 16.45       & 91.16       & 16.98        & 90.95        \\
\multicolumn{2}{c|}{SSDG-R \cite{DBLP:conf/cvpr/JiaZSC20}} & 7.38        & 97.17       & 10.44       & 95.94       & 11.71       & 96.59       & 15.61       & 91.54       & 11.29        & 95.31        \\
\multicolumn{2}{c|}{D$^2$AM \cite{DBLP:conf/aaai/ChenYSDTLHJ21}} & 12.70 & 95.66 &  20.98 & 85.58 & 15.43 & 91.22 & 15.27 & 90.87 & 16.10 & 90.83 \\
\multicolumn{2}{c|}{ViT  \cite{DBLP:conf/eccv/HuangSLCXYAY22}} & \textbf{4.75}        & \underline{98.59}       & 15.70       & 92.76       & 17.68       & 86.66       & 16.46       & 90.37       & 13.65        & 92.10        \\
\multicolumn{2}{c|}{TransFAS \cite{DBLP:journals/tbbis/WangWDG22}}    & 7.08        & 96.69       & 9.81        & 96.13       & 10.12       & 95.53       & 15.52       & 91.10       & 10.63        & 94.86        \\
\multicolumn{2}{c|}{DADN-CDS \cite{DBLP:journals/tcsv/YanZH22}}  & \underline{5.24}        & 98.06       & 6.84        & 97.95       & 10.64       & 95.14       & 13.77       & 93.09       & \textbf{9.12}         & \underline{96.06}        \\
\multicolumn{2}{c|}{CIFAS \cite{DBLP:conf/icmcs/LiuCDLZX22}}        & 5.95        & 96.32       & 10.66       & 95.30       & \textbf{8.50}        & \textbf{97.24}       & 13.17       & 93.44       & \underline{9.57}         & 95.58        \\
\multicolumn{2}{c|}{CF-PAD \cite{fang2024face}}             & 8.11        & 96.43       & 11.78       & 95.64       & 16.50       & 91.50      & 9.87        & 95.13       & 11.57        & 94.68       \\ \hline \hline
\multirow{3}{*}{ViT-B} 
& ViT-FS             & 11.19 & 96.09  & 9.89  & 95.67  & 14.90  &  93.59 & \textbf{5.52}  & \textbf{98.60}  & 10.37  & 95.99   \\
& FE                 & 30.71 & 77.50  & 18.67  & 90.33  & 36.10  & 72.71  & 27.07  & 80.74  & 28.14  & 80.32   \\
& FoundPAD (ours)    & 20.95 &  89.88 &  \textbf{4.89} & \underline{98.08}  & 10.45  & 95.80  & 6.19  & 98.31  & 10.62  & 95.52   \\ \hline
\multirow{3}{*}{ViT-L} 
& ViT-FS             & 8.10 &  98.12 & 26.11  &  82.97 &  21.55 & 87.29  & 36.40  & 69.57  & 23.04  & 84.49   \\
& FE                 & 21.67 &  86.87 & 9.00  & 96.10  & 22.05  & 86.27  & 22.32  & 84.85  & 18.76  & 88.52   \\
& FoundPAD (ours)    & 16.90 & 93.18  & \underline{6.00} & \textbf{98.72}  & \underline{9.90}  & \underline{96.07}  & \underline{5.87}  & \underline{98.41}  & 9.67  & \textbf{96.60}   \\ \hline
\end{tabular}}
\label{tab:sota_multiple_cross_db}
\end{center}
\vspace{-3mm}
\end{table*}

\begin{table}[]
\tiny
\begin{center}
\vspace{-3mm}
\caption{Results of triple-source cross-dataset evaluation on the CA dataset. The best and second-best results for each metric are highlighted in bold and underlined, respectively. FoundPAD surpasses both ViT-FS and FE in all considered scenarios, while achieving superior performances than SOTA methods for ViT-B.}
\vspace{-3mm}
\resizebox{0.41\textwidth}{!}{
\begin{tabular}{ll|cc}
\hline 
\multicolumn{2}{c|}{\multirow{2}{*}{Method}} & \multicolumn{2}{c}{O\&C\&M → CA} \\
 & & HTER(\%) $\downarrow$ & AUC(\%) $\uparrow$  \\ \hline \hline
\multicolumn{2}{c|}{GRL Layer \cite{DBLP:conf/icml/GaninL15}}       & 29.1 & 76.4 \\ 
\multicolumn{2}{c|}{ADDA \cite{DBLP:conf/cvpr/TzengHSD17}} & 33.7 & 70.3 \\ 
\multicolumn{2}{c|}{DA-FAS \cite{DBLP:conf/cvpr/SahaXKGCPG20}}       & 27.1 & 79.2 \\ 
\multicolumn{2}{c|}{UCDA-FAS \cite{DBLP:conf/fgr/PanwarSSPG21}}          & 26.1 & 80.0 \\ 
\multicolumn{2}{c|}{CIFAS \cite{DBLP:conf/icmcs/LiuCDLZX22}}            & 24.6 & 83.2 \\
% Baseline & 27.1 & 80.3 \\ 
\multicolumn{2}{c|}{CF-PAD \cite{fang2024face}}    & 23.5 & 84.2 \\  \hline \hline
\multirow{3}{*}{ViT-B} 
& ViT-FS            &  \underline{16.0} & \underline{89.1}  \\
& FE                &  23.7 & 84.2  \\
& FoundPAD (ours)   &  \textbf{15.6} & \textbf{91.0}  \\ \hline
\multirow{3}{*}{ViT-L} 
& ViT-FS            & 48.2  & 52.3  \\
& FE                & 43.9  & 58.3  \\
& FoundPAD (ours)   & 43.0 & 59.7  \\ \hline
\end{tabular}}
\label{tab:ocm_ca}
\vspace{-5mm}
\end{center}
\end{table}

\begin{table}[]
\tiny
\vspace{-3mm}
\begin{center}
\caption{Results of double-source cross-dataset evaluation. The best and second-best results for each metric are highlighted in bold and underlined, respectively. FoundPAD surpasses both ViT-FS and FE in all the considered scenarios while achieving superior performances than SOTA PAD methods.
}
\vspace{-3mm}
\resizebox{0.47\textwidth}{!}{
\begin{tabular}{ll|cc|cc}
\hline
\multicolumn{2}{c|}{\multirow{2}{*}{Method}} & \multicolumn{2}{c|}{M\&I → C} & \multicolumn{2}{c}{M\&I → O} \\
 & & HTER(\%) $\downarrow$ & AUC(\%) $\uparrow$ & HTER(\%) $\downarrow$ & AUC(\%) $\uparrow$     \\
\hline \hline
\multicolumn{2}{c|}{MS-LBP \cite{DBLP:conf/icb/MaattaHP11}}               & 51.16        & 52.09       & 43.63        & 58.07       \\
\multicolumn{2}{c|}{IDA \cite{msu_mfs}}                    & 45.16        & 58.80       & 54.52        & 42.17       \\
\multicolumn{2}{c|}{MADDG \cite{DBLP:conf/cvpr/ShaoLLY19}}                & 41.02        & 64.33       & 39.35        & 65.10       \\
\multicolumn{2}{c|}{RFM \cite{DBLP:conf/aaai/ShaoLY20}}                      & 36.34        & 67.52       & 29.12        & 72.61       \\
\multicolumn{2}{c|}{SSDG-R  \cite{DBLP:conf/cvpr/JiaZSC20}}   & 31.89        & 71.29       & 36.01        & 66.88       \\
\multicolumn{2}{c|}{DR-MD-Net \cite{DBLP:conf/cvpr/Wang0SC20}}  & 31.67        & 75.23       & 34.02        & 72.65       \\
\multicolumn{2}{c|}{D$^2$AM \cite{DBLP:conf/aaai/ChenYSDTLHJ21}} & 32.65 & 72.04 & 27.70 & 75.36  \\
\multicolumn{2}{c|}{CIFAS \cite{DBLP:conf/icmcs/LiuCDLZX22}}  & 22.67        & 83.39       & 24.63        & 81.48       \\ 
\multicolumn{2}{c|}{CF-PAD \cite{fang2024face}}                   & 22.11        & 85.06       & 19.71        & 89.01  \\ \hline \hline
\multirow{3}{*}{ViT-B} 
& ViT-FS            &  14.00 & 92.97  & \textbf{7.11}  & \textbf{97.88}  \\ 
& FE                & 27.22  & 79.94  & 33.57  & 72.70  \\ 
& FoundPAD (ours)   &  13.22 & 93.97  & 9.31  & \underline{96.69}  \\  \hline
\multirow{3}{*}{ViT-L} 
& ViT-FS            & 25.22  &  85.66 & \underline{9.07}  & 96.41  \\ 
& FE                &  \underline{11.33} & \underline{94.57} & 26.19  & 81.31  \\ 
& FoundPAD (ours)   & \textbf{4.67}  & \textbf{99.22}  & 10.23  & 95.58  \\  \hline
\end{tabular}}
\label{tab:mi_c_o}
\end{center}
\vspace{-3mm}
\end{table}

\begin{table*}[]
\tiny
\begin{center}
\vspace{-3mm}
\caption{Results of single-source cross-dataset evaluation. The best and second-best results for each metric are highlighted in bold and underlined, respectively. FoundPAD surpasses both ViT-FS and FE in most of the considered scenarios while achieving superior average performances than SOTA PAD methods by a large margin.}
\vspace{-3mm}
\resizebox{0.98\textwidth}{!}{
\begin{tabular}{ll|ccc|ccc|ccc|ccc||cc}
\hline
\multicolumn{2}{c|}{Method} & C → I   & C → M   & C → O  & I → C   & I → M   & I → O  & M → C   & M → I   & M → O   & O → I   & O → M   & O→  C   & Average  & Worst \\ \hline \hline
\multicolumn{2}{c|}{Binary CNN \cite{DBLP:journals/corr/YangLL14}} & 45.80 & 25.60  & 36.40 & 44.40 & 48.60 & 45.40 & 50.10 & 49.90  & 31.40 & 47.40 & 30.20 & 41.20  & 41.37 ± 8.42 & 50.10 \\ 
\multicolumn{2}{c|}{ADA \cite{DBLP:conf/icb/WangHSC19}}     & 17.50 & 9.30  & 29.10 & 41.60 & 30.50 & 39.60 & 17.70 & \underline{5.10} & 31.20  & 26.80 & 31.50 & 19.80   & 24.98 ± 11.28 & 41.60 \\ 
\multicolumn{2}{c|}{DR-MD-Net \cite{DBLP:conf/cvpr/Wang0SC20}} & 26.10 & 20.20 & 24.70 & 39.20 & 23.20 & 33.60  & 34.30 & 8.70 & 31.70  & 27.60 & 22.00 & 21.80   & 26.09 ± 7.70 & 39.20  \\ 
\multicolumn{2}{c|}{DR-UDA \cite{DBLP:journals/tifs/WangHSC21}}     & \underline{15.60} & 9.00 & 28.70 & 34.20 & 29.00 & 38.50  & 16.80 & \textbf{3.00} & 30.20 & 25.40 & 27.40 & 19.50 & 23.11 ± 10.50 & {38.50} \\ 
\multicolumn{2}{c|}{CF-PAD \cite{fang2024face}}               & 24.80 & 17.14 & 19.43 & 34.00 & 24.76 & 31.70  & 14.44 & 15.90 & 25.34 & 21.50 & \underline{15.00} & 20.33  & 22.03 ± 6.33 & 34.00 \\ \hline \hline
\multirow{3}{*}{ViT-B} 
& ViT-FS            & 26.05  & \underline{8.33}  & 17.79  & 24.33  & 22.38  & 18.43  & \textbf{4.00}  & 15.05  & \textbf{6.68}  & 20.45  & 15.48  & 11.56  & \underline{15.88 ± 7.07}  & 26.05  \\ 
& FE                &  32.95 & 35.24  & 30.63  & 38.56  & 35.71  & 40.52  & 25.33  & 30.45  & 28.86  & 34.65  & 32.86  & 16.89  & 31.89 ± 6.29   &  40.52 \\ 
& FoundPAD (ours)   &  16.40 & 24.52  & \underline{15.14}  & \underline{17.00}  & \textbf{18.57}  & \textbf{13.38}  & 20.00  & 17.10  & \underline{19.41}  & \textbf{8.95}  & 23.33  & \underline{7.89}  & 16.81 ± 5.03   & \underline{24.52}  \\  \hline
\multirow{3}{*}{ViT-L} 
& ViT-FS            & 22.60  & \textbf{5.71}  & 37.07  & 25.00  & 20.48  & 23.62  & \underline{7.89}  & 15.00  & 22.93  & 26.05  & \textbf{11.43}  & 20.33  & 19.84 ± 8.69  & 37.07  \\ 
& FE                & 25.70  &  30.24 & 25.06  & 19.78  & 26.90  & 31.83  & 15.44  & 19.00  &  28.37 & 24.40  & 32.62  & 13.78  & 24.43 ± 6.21 &  32.62 \\ 
& FoundPAD (ours)   &  \textbf{14.05} & 21.43  & \textbf{11.00}  & \textbf{10.22}  & \underline{19.29}  & \underline{16.94}  & 12.00  & 14.55  & 20.93  & \underline{14.40}  & 23.81  &  \textbf{7.22} & \textbf{15.49 ± 5.07}  & \textbf{23.81}  \\  \hline
\end{tabular}}
\label{tab:singe-db-results}
\vspace{-3mm}
\end{center}
\end{table*}

\begin{table*}[tbh!]
\tiny
\centering
\vspace{-3mm}
\caption{Results of single-source cross-dataset evaluation when training on the synthetic dataset SynthASpoof for previously proposed solutions and our proposed method, FoundPAD. The best and second-best results for each metric are highlighted in bold and underlined, respectively. FoundPAD presents the best average AUC and second-best average HTER, highlighting its generalizability to unseen domains.}
\vspace{-3mm}
\resizebox{0.99\textwidth}{!}{
\begin{tabular}{ll|cc|cc|cc|cc||cc}
\hline
\multicolumn{2}{c|}{\multirow{2}{*}{Method}}     & \multicolumn{2}{c|}{M} & \multicolumn{2}{c|}{C} & \multicolumn{2}{c|}{I} & \multicolumn{2}{c||}{O} & \multicolumn{2}{c}{Average} \\ %\cline{2-11}
\multicolumn{2}{c|}{}& HTER(\%) $\downarrow$ & AUC(\%) $\uparrow$   & HTER(\%) $\downarrow$ & AUC(\%) $\uparrow$  & HTER(\%) $\downarrow$ & AUC(\%) $\uparrow$    & HTER(\%) $\downarrow$ & AUC(\%) $\uparrow$   & HTER(\%) $\downarrow$ & AUC(\%) $\uparrow$      \\ \hline \hline
\multicolumn{2}{c|}{ResNet \cite{DBLP:conf/cvpr/FangHD23}}                 & 25.48             &  79.54                & 39.22     & 62.00  & 8.90               & \underline{96.96}   & 34.23             & 71.48             & 26.96             & 77.50 \\
\multicolumn{2}{c|}{PixBis \cite{DBLP:conf/cvpr/FangHD23}}                 & 38.33             &  63.87                & 38.44     & 64.79  & \underline{7.50}   & 96.88               & 35.77             & 63.50             & 30.74             & 72.26 \\
\multicolumn{2}{c|}{ViT-SIDE B  \cite{DBLP:conf/icb/FangHFRDAKPYHCZPJLSWLCZTSAS23}}            & 36.67             &  69.78                & 33.33     & 75.21  & 9.80               & 96.67               & \textbf{13.26}    & \textbf{94.04}    & \textbf{23.27}    & \underline{83.93} \\
\multicolumn{2}{c|}{SynFace Co-Former A  \cite{DBLP:conf/icb/FangHFRDAKPYHCZPJLSWLCZTSAS23}}   & \underline{18.57} &  \underline{90.76}    & 41.11     & 64.49  & 16.30              & 92.31               & \underline{21.67} & \underline{86.44} & 24.41             & 83.50 \\
\multicolumn{2}{c|}{SynFace Co-Former B \cite{DBLP:conf/icb/FangHFRDAKPYHCZPJLSWLCZTSAS23}}    & \textbf{16.67}    &  \textbf{91.61}       & 40.00     & 63.05  & 18.80              & 88.20               & 25.35             & 82.02             & 25.21             & 81.22 \\
\multicolumn{2}{c|}{CoDe-Lc \cite{DBLP:conf/icb/FangHFRDAKPYHCZPJLSWLCZTSAS23}}                & 37.14             &  71.45                & 37.11     & 69.08  & 12.10              & 95.31               & 37.58             & 66.30             & 30.98             & 75.54 \\
\multicolumn{2}{c|}{CoDe-Lh  \cite{DBLP:conf/icb/FangHFRDAKPYHCZPJLSWLCZTSAS23}}               & 39.05             &  70.58                & 39.33     & 63.70  & 13.90              & 93.84               & 38.11             & 68.33             & 32.60             & 74.11 \\
\multicolumn{2}{c|}{OrthPADNet \cite{DBLP:conf/icb/FangHFRDAKPYHCZPJLSWLCZTSAS23}}             & 20.95             &  87.59                & 39.78     & 67.32  & 23.70              & 79.55               & 34.92             & 71.69             & 29.84             & 76.54 \\
\multicolumn{2}{c|}{idvcVT \cite{DBLP:conf/icb/FangHFRDAKPYHCZPJLSWLCZTSAS23}}                 & 45.71             &  64.58                & 56.44     & 41.82  & 23.10              & 85.08               & 51.09             & 49.68             & 44.09             & 60.29 \\
\multicolumn{2}{c|}{hdaFVPAD \cite{DBLP:conf/icb/FangHFRDAKPYHCZPJLSWLCZTSAS23}}               & 65.71             &  31.95                & 71.33     & 21.95  & 47.80              & 52.10               & 37.89             & 66.49             & 55.68             & 43.12 \\
\hline \hline
\multirow{3}{*}{ViT-B} 
& ViT-FS                     & 50.24 & 58.61  & 44.44  & 59.46  & 24.40  & 81.48  & 46.53  & 56.08  & 41.40  & 63.91  \\
& FE                         & 47.14 & 59.27  & 28.11  & 78.81  & 19.50  & 87.08  & 40.28  & 63.66  & 33.76  & 72.21  \\
& FoundPAD (ours)            & 47.14 & 66.18  & 27.33  & 83.03  & 16.15  & 90.79  & 33.12  & 73.56  & 30.94  & 78.39  \\ \hline
\multirow{3}{*}{ViT-L} 
& ViT-FS                     & 50.00 & 55.94  & 47.11  & 58.14  & 33.60  & 73.20  & 50.04  & 50.63  & 45.19  & 59.48  \\
& FE                         & 52.62 & 55.76  & \underline{13.89}  & \underline{92.82}  &  20.50 &  87.94 &  29.58 &  77.13 & 29.15  & 78.41  \\
& FoundPAD (ours)            & 45.71 & 69.76  & \textbf{9.89}  & \textbf{96.03}  & \textbf{6.40}  & \textbf{98.58}  & 32.05  & 75.69  & \underline{23.51}  & \textbf{85.01}  \\  \hline
\end{tabular}}
\vspace{-3mm}
\label{tab:synthaspoof}
\end{table*}

\vspace{-2mm}
\paragraph{Zero-Shot PAD (TI):} As described in Section \ref{sec:baseline}, each test sample was paired with textual descriptions of ``biometric presentation attack'' and ``bona-fide presentation'', with the similarity score between the image embedding and the text embeddings determining the predicted label. The results of this evaluation are presented in Table \ref{tab:no_training}. The results reveal CLIP's limitations for PAD detection without training, with high HTERs and low AUCs, performing near random. ViT-L shows slight improvement over ViT-B, but the overall performance remains poor. This is likely due to CLIP's text encoder, which lacks domain-specific semantics, struggles with nuanced PAD characteristics, and fails to align effectively with dataset features. These results highlight the need for domain-adapted fine-tuning or enhanced prompts for effective PAD using CLIP.

\vspace{-4mm}
\paragraph{Baselines toward FoundPAD (ViT-FS and FE):} Since using image-text pairs did not achieve any tangible results, we further explored the embedding space of the image encoder of the FM, which is induced by the pre-trained weights of the model. To measure this quantitatively, two different approaches have been used as described in Section \ref{sec:baseline}: ViT-FS and FE. To this end, ViT-FS and FE are evaluated under triple-source, double-source, and single-source scenarios with two different architectures, whose results are depicted in Tables \ref{tab:sota_multiple_cross_db}, \ref{tab:mi_c_o}, and \ref{tab:singe-db-results} respectively. With a higher availability of the data, i.e. the triple-source case, ViT-FS achieved better results for all cases than FE for ViT-B, whereas ViT-FS performed worse than FE for ViT-L. Given the reduced size of the datasets used to train ViT-FS and FE, ViT-FS is likely to underperform when a larger architecture is considered, as it does not benefit from the FM's previous knowledge. On the other hand, the number of trainable parameters in the ViT-B network is significantly smaller, which might justify ViT-FS's increased performance in this scenario. For ViT-B, ViT-FS had an average HTER of 10.37\% and an AUC of 95.99\%, whereas FE achieved an average HTER of 28.14\% and an AUC of 80.32\% as can be seen in Table \ref{tab:sota_multiple_cross_db}. In lower data availability settings, i.e. the double-source case or the single-source case, the performance gap between ViT-FS and FE remains large. The double-source case in Table \ref{tab:mi_c_o} shows ViT-FS achieving an HTER of 14.00\% and 7.11\% on CASIA-FASD and OULU-NPU, respectively, while FE achieves 27.22\% and 33.57\% for these metrics, when considering ViT-B. The same phenomenon is seen for the single-source case in Table \ref{tab:singe-db-results} as HTER averages are 15.88\% for ViT-FS and 31.89\% for FE with the ViT-B. Similar gaps are also observed for ViT-L for the single-source case. Hence, while using the embedding space of the FM is a good start and can achieve better results than a binary CNN \cite{DBLP:journals/corr/YangLL14}, it is by no means an optimal embedding space. This suggests that either the embedding space of the FM should be aligned to the PAD task or a deeper classification network should be used.

\vspace{-4mm}
\paragraph{FoundPAD:} Since the FMs' embeddings should be aligned to the PAD, the exploration of adapting the embedding space for PAD-specific nuances leads to the FoundPAD framework described in Section \ref{sec:methodology}. Toward this goal, FoundPAD has been evaluated on the same three scenarios as for ViT-FS and FE. The comparison with ViT-FS allows us to withdraw conclusions regarding the power of FMs' pre-training on large-scale databases and further adaption to the available data. On the other hand, the comparison between FoundPAD and FE shows the re-usability of the original FM's embedding space and the benefits of adapting the FM with LoRA to the PAD downstream task. For the triple-source scenario (Table \ref{tab:sota_multiple_cross_db}), FoundPAD achieved an average HTER of 10.62\% and an AUC of 95.52\% for ViT-B, and an average HTER of 9.67\% and an AUC of 96.60\% for ViT-L. In comparison to the results for ViT-FS and FE, this shows an AUC increase of 13.37 pp. compared to ViT-FS and 9.09 pp. compared to FE with ViT-L. Likewise, an increase in AUC was observed with ViT-B, specifically 17.52 pp. to FE, and FoundPAD achieved similar averages to ViT-FS. For the double-source scenario (Table \ref{tab:mi_c_o}), FoundPAD achieved high AUCs and low HTERs evaluated on either CASIA-FASD or OULU-NPU for both architectures.

For the single-source PAD (Table \ref{tab:singe-db-results}), FoundPAD was able to surpass ViT-FS in 7 out of the 12 evaluated scenarios with ViT-B, and in 9 out of the 12 evaluated scenarios with ViT-L, whereas FoundPAD outperformed FE in all evaluated scenarios regardless of the architecture. The comparison between FoundPAD and FE showed improved performance in all the analyzed scenarios for triple, double and single-source settings, leading to the conclusion that adapting the embedding space for PAD-specific nuances and aligning the embedding space to PAD was necessary. Furthermore, FoundPAD performed very competitively to ViT-FS, highlighting the benefits of taking advantage of the built-in knowledge of the pre-trained FM.

\vspace{-5mm}
\paragraph{Comparison with SOTA:} As PAD research is always evolving and new approaches have been proposed rapidly in the last years, we did our best to provide the most comprehensive comparison with the recent works across all experimental setups, however, we acknowledge that the comparison might have missed specific works and acts as a tool to place the achieved performances within the scope or SOTA, rather than a comprehensive comparison. We also state that we list the works that evaluated the corresponding protocols in each table, so the approaches in each table might not completely overlap. In this part, we will exclude ViT-FS and FE comparisons and only compare with FoundPAD. \textbf{Triple-source cross-dataset evaluation:} Comparison with the SOTAs are gathered in Tables \ref{tab:sota_multiple_cross_db}, and \ref{tab:ocm_ca} as different protocols were followed by the different methods. Table \ref{tab:sota_multiple_cross_db} shows that the best average AUC is achieved by FoundPAD-ViT-L (96.60\%), followed by TransFAS \cite{DBLP:journals/tbbis/WangWDG22} (96.06\%). For the evaluations on C and O, FoundPAD-ViT-B and FoundPAD-ViT-L share 1st and 2nd places. Evaluation on I shows competitive results with others, whereas evaluation on M is where FoundPAD needs improvement. In Table \ref{tab:ocm_ca}, we see that FoundPAD-ViT-B achieves improved results, whereas FoundPAD-ViT-L is not performing on the same level. \textbf{Double-source cross-dataset evaluation:}
From Table \ref{tab:mi_c_o}, we observe that FoundPAD-ViT-B and FoundPAD-ViT-L are the only methods that surpass the 90\% AUC mark, where FoundPAD-ViT-L reaches an AUC of 99.22\% on C and FoundPAD-ViT-B reaches an AUC of 96.69\% on O. \textbf{Single-source cross-dataset evaluation:} This setting represents the challenging low data availability scenario and is detailed in Table \ref{tab:singe-db-results}.
FoundPAD-ViT-B improved the best average HTER by 5.22 pp. while reducing the lowest worst achieved HTER from 34.00\% to 24.52\%. FoundPAD-ViT-L improved the best average HTER by 6.54 pp. while further reducing the lowest worst achieved HTER to 23.81\%. This illustrates FoundPADs' ability to use its induced knowledge from the FM.

\vspace{-5.6mm}
\paragraph{Synthetic Data Applicability for FoundPAD:}
To showcase that the proposed FoundPAD can also be fine-tuned using privacy-friendly synthetic data, we utilize SynthASpoof \cite{DBLP:conf/cvpr/FangHD23}. Not only do we compare with \cite{DBLP:conf/cvpr/FangHD23} but also with all methods presented to the SynFacePAD 2023 competition \cite{DBLP:conf/icb/FangHFRDAKPYHCZPJLSWLCZTSAS23}. We obtained comparable results to the winner of the competition ``ViT-SIDE B'', which proposed a ViT model architecture pre-trained on ImageNet \cite{dosovitskiy2021imageworth16x16words}. We also note that our conclusions on comparing ViT-FS, FE, and FoundPAD still stand when using synthetic data.

\vspace{-2mm}
\section{Conclusion} \label{sec:conclusion}
\vspace{-2mm}
In this work, we propose the first PAD method that takes advantage of FMs potential to perform generalizable classification even with low data availability, FoundPAD. In particular, we adapt the pre-trained FM CLIP to the PAD task using low-rank adaption (LoRA), while simultaneously training a header to perform classification. This allows the FM to adapt its feature space to the downstream PAD task without losing the knowledge acquired during its self-supervised pre-training process. To show that FoundPAD is effectively taking advantage of FMs' properties, we further propose and evaluate three alternative FM and transformer-based methods, TI, ViT-FS and FE, whose performances we compare with FoundPAD on different data availability settings. These experiments proved the overall superiority of FoundPAD compared with the remaining approaches, highlighting that the in-built knowledge of the FM is beneficial for the downstream task while not being enough to achieve good classification performances in domain-specific tasks such as PAD. FoundPAD also achieved competitive results or even surpassed PAD SOTA methods.
Further experiments using synthetic training data also demonstrate FoundPAD's ability to generalize to the unseen authentic domain on several evaluation benchmarks. These outcomes show the feasibility of using FoundPAD to tackle common PAD issues due to its high generalization capacity, namely in the challenging low data availability scenario.

\textbf{Acknowledgment:} This research work has been funded by the German Federal Ministry of Education and Research and the Hessian Ministry of Higher Education, Research, Science and the Arts within their joint support of the National Research Center for Applied Cybersecurity ATHENE.

{\small
\bibliographystyle{ieee_fullname}
\bibliography{egbib}
}

\end{document}